\pdfoutput=1

\documentclass[11pt]{article}

\usepackage[]{acl}

\usepackage{times}
\usepackage{latexsym}

\usepackage{graphicx}
\usepackage{xcolor}
\usepackage{multirow}
\usepackage{hhline}
\usepackage{array}
\usepackage{hyperref}
\usepackage{CJK}

\usepackage{amsmath}
\usepackage{amssymb}
\usepackage{subcaption}

\usepackage{listings}

\usepackage{color}
\definecolor{codegreen}{rgb}{0,0.6,0}
\definecolor{codegray}{rgb}{0.5,0.5,0.5}
\definecolor{codepurple}{rgb}{0.58,0,0.82}
\definecolor{backcolour}{rgb}{0.95,0.95,0.92}

\lstdefinestyle{mystyle}{
    backgroundcolor=\color{backcolour},   
    commentstyle=\color{codegreen},
    keywordstyle=\color{magenta},
    numberstyle=\tiny\color{codegray},
    stringstyle=\color{codepurple},
    basicstyle=\footnotesize,
    breakatwhitespace=false,         
    breaklines=true,                 
    captionpos=b,                    
    keepspaces=true,                                    
    numbersep=5pt,                  
    showspaces=false,                
    showstringspaces=false,
    showtabs=false,                  
    tabsize=2
}

\usepackage[T1]{fontenc}

\usepackage[utf8]{inputenc}

\usepackage{microtype}

\usepackage{inconsolata}

\title{ICE-GRT: Instruction Context Enhancement by Generative Reinforcement based Transformers}

\author{\parbox{0.9\linewidth}{
\centering{Chen Zheng  ~ Ke Sun ~ Da Tang ~ Yukun Ma ~ Yuyu Zhang \\ Chenguang Xi ~ Xun Zhou 
} \\
{\rm AML Group, Bytedance Inc.} \\
\texttt{\{chen.zheng1,ke.sun1,da.tang,mayukun,yuyu.zhang\}@bytedance.com} \\
\texttt{\{chenguang.xi,zhouxun\}@bytedance.com} \\
}
}

\begin{document}
\maketitle
\begin{abstract}

The emergence of Large Language Models (LLMs) such as ChatGPT and LLaMA encounter limitations in domain-specific tasks, with these models often lacking depth and accuracy in specialized areas, and exhibiting a decrease in general capabilities when fine-tuned, particularly analysis ability in small sized models. To address these gaps, we introduce ICE-GRT, utilizing Reinforcement Learning from Human Feedback (RLHF) grounded in Proximal Policy Optimization (PPO), demonstrating remarkable ability in in-domain scenarios without compromising general task performance. Our exploration of ICE-GRT highlights its understanding and reasoning ability to not only generate robust answers but also to provide detailed analyses of the reasons behind the answer. This capability marks a significant progression beyond the scope of Supervised Fine-Tuning models. The success of ICE-GRT is dependent on several crucial factors, including Appropriate Data, Reward Size Scaling, KL-Control, Advantage Normalization, etc. The ICE-GRT model exhibits state-of-the-art performance in domain-specific tasks and across 12 general Language tasks against equivalent size and even larger size LLMs, highlighting the effectiveness of our approach. We provide a comprehensive analysis of the ICE-GRT, underscoring the significant advancements it brings to the field of LLM.

\end{abstract}

\section{Introduction}
The advent of Large Language Models (LLMs) like ChatGPT~\cite{Brown2020LanguageMA,OpenAI2023GPT4TR} and LLaMA~\cite{Touvron2023LLaMAOA,Touvron2023Llama2O} has marked a significant milestone in the field of Natural Language Processing (NLP). These models have gained widespread recognition for their robust general conversational abilities, enabling fluid and coherent responses across a diverse range of topics. However, there are key limitations to these models. 
Firstly, a key limitation surfaces when these models encounter domain-specific tasks~\cite{Zhao2023ASO,zhang2023balancing}. In scenarios that demand deep technical knowledge or specialized expertise, these models often fall short, providing responses that lack necessary depth and accuracy. Secondly, Supervised Fine Tune (SFT) LLMs tend to exhibit a decrease in general capabilities~\cite{Ling2023DomainSA}. This is contrary to the expectations held for large-scale models, which are presumed to either maintain or improve their performance in a wide array of tasks~\cite{Pan2023APE}. Lastly, the current smaller-sized LLMs, such as $13$ Billion, demonstrate a limited ability to conduct detailed analysis on complex questions, a competency that is significantly inferior compared to the capabilities of models like ChatGPT, which can engage in more comprehensive and detailed discussions.

Addressing these challenges, we introduce the \textbf{I}nstruction \textbf{C}ontext \textbf{E}nhancement by \textbf{G}enerative \textbf{R}einforcement based \textbf{T}ransformers~(ICE-GRT), an innovative LLM that leverages the principles of Reinforcement Learning from Human Feedback~(RLHF)~\cite{Brown2020LanguageMA} based on Proximal Policy Optimization (PPO)~\cite{schulman2017proximal}. While ensuring that the general capabilities of the Large Language Model (LLM) are maintained, ICE-GRT exhibits exceptional performance in several domain-specific scenarios. 
Furthermore, ICE-GRT demonstrates an improved ability for detailed analysis, particularly in complex scenarios where smaller-sized LLMs fall short.

We take one domain-specific task of ad moderation as an example. ICE-GRT can not only determine the compliance of advertisements but also identify the specific category of violation. Moreover, it goes a step further by detailed analyzing which elements of the ad are problematic and offers constructive modification suggestions. This is a notable advancement over both pretrained and SFT~\cite{vicuna2023} LLM models, which are typically limited to identifying compliance and violation categories.

When our training methodology was applied to RLHF, we observed not just significant improvements in in-domain tasks but also a surprising enhancement in general tasks.
In a comparative analysis against models of equivalent and larger parameter size across many general tasks, our ICE-GRT model with 13 billion parameters consistently achieved state-of-the-art performance in $12$ well-known public LLM evaluation benchmarks.
ICE-GRT's versatility is further illuminated through its effective handling of various domain-specific tasks, not limited to but including Poem Generation, Text-to-Table conversions, engaging Multiple Round Dialogue, generating accurate Multi-lingual Responses, proficient Code Generation, creating tailored Ads Text and Labeling Text, etc.

Our exploration of the ICE-GRT model has uncovered several factors critical to its training success. The ICE-GRT model's training data, sourced from our ICE-Instruct (SFT) model and enriched with human feedback with strict evaluation criteria, offers a diverse and comprehensive dataset, essential for its robust training. Moreover, the scaling of the reward model is essential for accurately capturing complex scenarios and aligning with human preferences in RLHF. Additionlly, KL-Control is key to regulating the balance between the models, while Advantage Normalization significantly improves learning stability by adjusting advantage estimates. Additionally, we discovered that modifying the Clipping Range and carefully controlling the maximum response length during sampling are vital for enhancing the training process. These findings deepen our understanding of RLHF mechanisms and are instrumental in effectively training the ICE-GRT model.

Moreover, we provide a detailed analysis of the ICE-GRT model, encompassing both general and in-domain capabilities. Through this exploration, we aim to contribute a novel perspective and methodology to the field of NLP, particularly in enhancing the depth and accuracy of domain-specific task handling by large language models.
We observe that the pretrain phase engages in ``knowledge learning'', where the model extensively absorbs a diverse range of information, forming a substantial foundational knowledge base. Subsequently, in the Supervised Fine-Tuning stage, the model engages in ``knowledge mining'', where it utilizes the learned knowledge in response to specific instructions. This stage is crucial for the model to transition from passive knowledge accumulation to active knowledge application. Finally, the RLHF phase engages in ``knowledge enhancement'', enhancing the model's ability to align with human language preferences. This stage builds upon the vast knowledge gained in the pretrain phase and the knowledge mining from the SFT stage, leading to a model that not only reconstruct extensive knowledge but also excels in applying it with human-centric preference. Importantly, this phase showcases a significant leap in the model's emergence capabilities.

In our commitment to fostering collaborative research and innovation, \textbf{we make ICE-GRT publicly available on HuggingFace}\footnote{Our ICE-GRT is
available at \url{https://huggingface.co/zhengchenphd/ICE-GRT}.}. This open-source initiative is aimed at empowering researchers globally to further investigate and expand upon our findings with ICE-GRT. By democratizing access to this advanced model, we hope to inspire and facilitate worldwide exploration and progress in language model research. 
This paper unveils just a fraction of ChatGPT's capabilities, and our choice of the acronym "ICE" for ICE-GRT is purposeful. It represents our aspiration to accelerate the 'ice-breaking' process in LLM research, symbolizing our desire to inspire researchers to explore and uncover the vast potential of ICE-GRT across an array of tasks and paving the way for new discoveries and advancements in the field.

\section{Related Works}


\subsection{Instruction-Tuning for LLM}
Recent advancements in Large Language Model (LLM) development have increasingly focused on instruction-tuning~\cite{vicuna2023}, a technique that is gaining significant traction particularly within the realms of     Question Answering (QA) and different domains~\cite{Zhao2023ASO,pan2023preliminary,qiu2020pre}. Key research in this area includes works such as ALPACA~\cite{alpaca}, Vicuna~\cite{vicuna2023}, and~\cite{Zhang2023BalancingSA}, which explores the balance between diveristy and accuracy in large language model. Furthermore, studies like~\cite{sun2023principle} delve into principles of effective QA strategies, while~\cite{zhou2023lima} present LIMA, an innovative model for language interaction. In the sphere of conversational interfaces, significant contributions include the development of OpenAssistant by~\cite{kopf2023openassistant, vicuna2023}.

\subsection{Reinforcement Learning from Human Feedback (RLHF)}
Alongside the development of LLMs, Reinforcement Learning from Human Feedback has emerged as an important approach to improve LLMs~\cite{Brown2020LanguageMA,Touvron2023Llama2O}. RLHF involves training models not just on static datasets but also incorporating human feedback to guide the learning process. This method has been particularly useful in aligning knowledge learning and mining with human feedback. For instance, models like OpenAI's InstructGPT have utilized RLHF to tailor responses based on human preferences, leading to more accurate outputs~\cite{stiennon2020learning}.

\section{Model}

\begin{figure}
\centering
\includegraphics[width=0.45\textwidth,height=160pt]{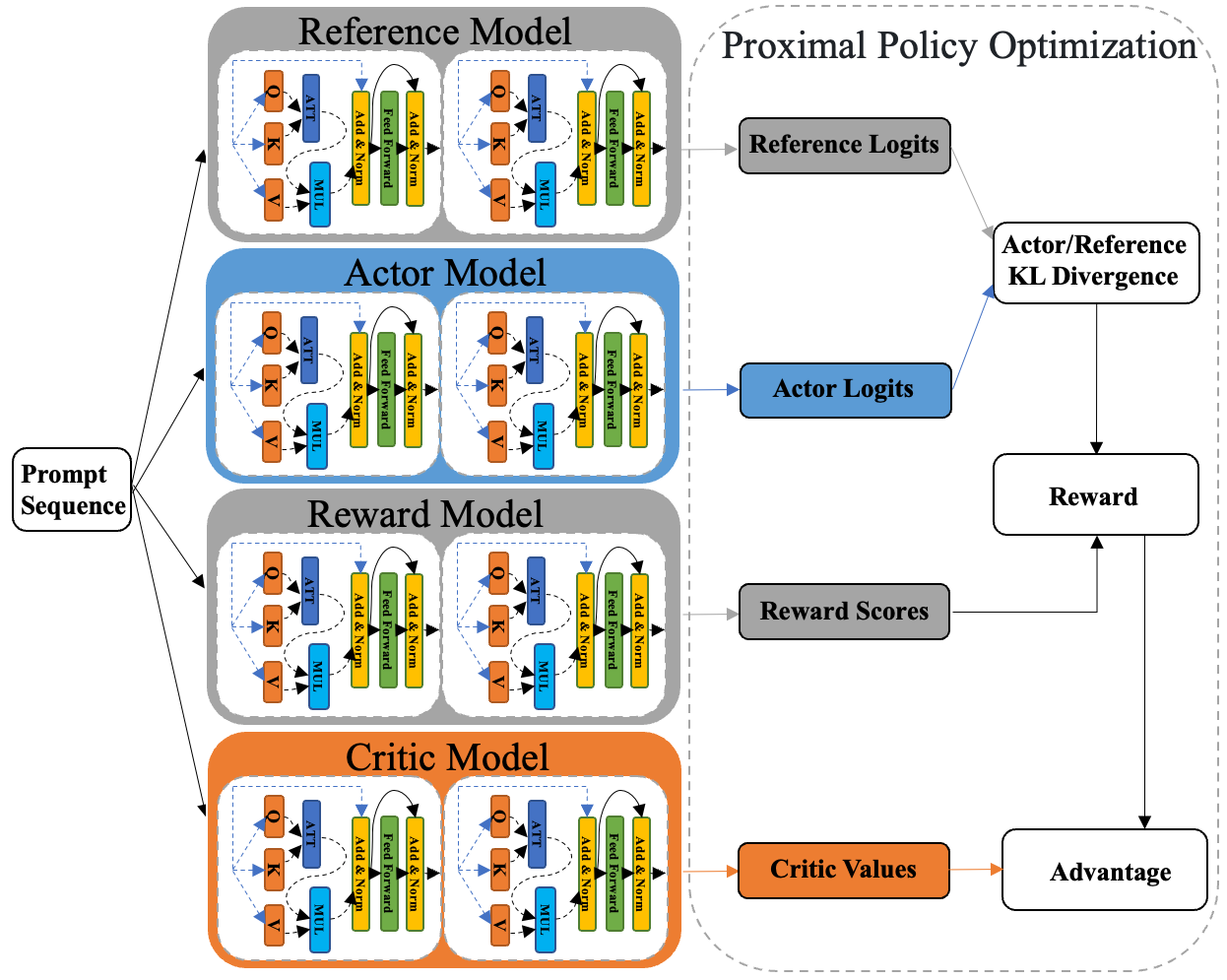}
\caption{ICE-GRT Model Architecture.\label{fig:icegrt_architecture}}
\end{figure}

In this section, we briefly introduce a SFT model we have trained, named ICE-Instruct, designed to improve the domain-specific knowledge mining capabilities of pre-trained LLMs. Following this, we will give a detailed description of our process for training the reward model, which we have termed ICE-Reward. Finally, we will comprehensively introduce the entire training process of ICE-GRT, including some important training strategies.

\subsection{ICE-Instruct}

The ICE-Instruct model built upon the Vicuna model~\cite{vicuna2023}. By blending in-domain and general-purpose data during fine-tuning, it excels in both specialized tasks and broader tasks. This approach not only maintains its vast linguistic capacities but also enhances its expertise in specific domains. 
Importantly, this sets a solid foundation for RLHF models. 
All subsequent actor and critic models are initialized using ICE-Instruct as backbone. In essence, ICE-Instruct determines the lower-bound capabilities of ICE-GRT, ensuring a strong and reliable baseline for further advancements. 
To maximize the model's applicability in contextual interactions, we have converted all collected data into Question-Answer pairs. 
Each data point adheres to a prompt format that begins with \textit{``Below is an instruction that describes a task. Write a response that appropriately completes the request. \#\#\# USER: <INPUT> ASSISTANT: <OUTPUT> ''}, ensuring consistency and relevance in contexts.

\subsection{ICE-Reward}
\noindent \textbf{Response Generation and Sampling: } Initially, for each prompt in the RLHF training dataset, we generate five responses. These responses are uniquely produced by our \textbf{ICE-Instruct} model. By sampling from the model's output distribution, we ensure a diverse range of generated answers
, capturing various aspects of potential responses.

\noindent \textbf{Human Annotation and Ranking: }
The generated responses are then subjected to human annotation. Annotators rank these responses according to predefined criteria detailed in section~\ref{sec:manual_evaluation}. 
Specifically, we labeled 20,000 sets of rankings, each set containing five responses.
From the ranked responses, we extract the top two and the bottom two responses for each prompt. These are then paired to form training data. The pairs consist of a ``better'' response and a ``worse'' response, as determined by the human annotation. This pairing strategy is instrumental in teaching the model the differences between high-quality and low-quality responses.

\noindent \textbf{Training Reward Model: }
The objective of training reward model is to develop a model capable of accurately differentiating between high and low-quality responses. Let \( R(s, a) \) be the reward function, where \( s \) represents the input prompt and \( a \) the generated response. Our goal is to optimize \( R \) so that it aligns with human judgments.
The training data consists of pairs \((a_i, a_j)\) where \(a_i\) is a higher-ranked response compared to \(a_j\) for the same prompt.
We use a pairwise ranking loss function, defined as:
$$ \mathcal{L}(a_i, a_j) = \max(0, \text{margin} - R(s, a_i) + R(s, a_j)). $$
This loss function encourages the model to assign a higher score to \(a_i\) than \(a_j\).

The trained reward model, therefore, learns to assign higher scores to more relevant and contextually appropriate responses, as per human rankings. This model forms a most critical part of our system, ensuring high-quality, context-aware responses.

\subsection{ICE-GRT}

In this section, we provide a comprehensive overview of each component involved in ICE-GRT, leverages the principles of RLHF~\cite{Brown2020LanguageMA} based on PPO~\cite{schulman2017proximal}, along with their respective mathematical formulations. Figure~\ref{fig:icegrt_architecture} shows the whole training process. 

\noindent \textbf{Actor Model:} 
The Actor model, represented as \(\pi_{\theta_\text{act}}(a|s)\), maps states \(s\) to actions \(a\). It is responsible for generating actor logits, which are scores assigned to each potential action.

\noindent \textbf{Reference Model:} 
The Reference model, denoted as \(\pi_{\theta_{\text{ref}}}(a|s)\), serves as a pre-trained benchmark for evaluating behavior. It provides a baseline against which the Actor model's outputs are compared throughout the training process.

\noindent \textbf{Reward Model:} 
The Reward model, expressed as \(R(s, a)\), assigns a reward score based on the quality of the generated sequence, evaluating both the action \(a\) and the state \(s\).

\noindent \textbf{Critic Model:} 
The Critic model, \(V_{\theta_\text{crt}}(s)\), estimates the value of being in a specific state \(s\), thereby producing critic values that guide the learning process.

\subsubsection{Generalized Advantage Estimation (GAE) Calculation in ICE-GRT}

The advantage function, \(A(s, a)\), assesses the relative benefit of executing a specific action in contrast to the average action in a given state. The formula for calculating the Advantage is:
\begin{equation}
A(s, a) = \mathbb{E}( R(s, a) + \gamma V_{\theta_\text{crt}}(s') - V_{\theta_\text{crt}}(s))
\end{equation}
where \(\gamma\) represents the discount factor, \(s'\) is the subsequent state following the current state \(s\), and $V_{\theta_\text{crt}}(s)$ is the value function estimated by the Critic model with weights $\theta_\text{crt}$.

Generalized Advantage Estimation (GAE), enhances the estimation of the advantage function in RL~\cite{Schulman2015HighDimensionalCC}. GAE blends multi-step return methods with value function estimates to mitigate variance while preserving a reasonable bias.
The essence of GAE is the employment of a weighted sum of n-step Temporal Difference (TD) residuals:
\begin{equation}
\delta^A_t = \mathbb{E}( R^{t+1}(s, a) + \gamma V_{\theta_\text{crt}}^{t+1}(s') - V_{\theta_\text{crt}}^{t}(s))
\end{equation}
Here, \(\delta^A_t\) represents the TD residual at time t.
Further, the GAE advantage function is calcuated as:
$A_\text{GAE}(s, a) = \sum_{l=0}^{\infty} (\gamma \lambda)^l \delta^A_{t+l}$,
where \(\lambda\) \(\in(0,1)\).

\subsubsection{Actor Model Learning}
The Actor Model is updated using the Proximal Policy Optimization objective~\cite{schulman2017proximal}, the process is calculated as follows:
\begin{equation} \small
    \begin{split}
    L(\theta_{\text{act}}) = \min \Bigg( & \frac{\pi_{\theta_{\text{act}}}(a|s)}{\pi_{\theta_{\text{old}}}(a|s)} A_\text{GAE}^{\pi_{\theta_{\text{old}}}}(s, a), \\
    & \text{clip}\left(\frac{\pi_{\theta_{\text{act}}}(a|s)}{\pi_{\theta_{\text{old}}}(a|s)}, 1 - \varepsilon, 1 + \varepsilon \right) A_\text{GAE}^{\pi_{\theta_{\text{old}}}}(s, a) \Bigg),
    \end{split}
\end{equation}
where \(A_\text{GAE}^{\pi_{\theta_{\text{old}}}}(s, a)\) is the advantage function calculated using the old policy \(\pi_{\theta_{\text{old}}}\), \(\varepsilon\in(0,1)\) is a hyperparameter. This term ensures that the evolving Actor policy remains not only stable in its updates but also aligned or divergent as desired from the old model.

\subsubsection{Policy Optimization and Training}
In the final stage, the PPO algorithm optimizes the Actor model's policy based on the calculated advantages, the KL-divergence, and the updated Actor model. The policy is iteratively updated to maximize the expected rewards, with the aim of aligning the Actor model's behavior more closely with established benchmarks while also ensuring effective and efficient learning.

\subsubsection{Important Training Strategies}

\noindent \textbf{ICE-GRT Training Data}:
Our ICE-GRT's training data originates from ICE-Instruct model and careful human feedback annotation. This data is not just a collection of responses but is intricately designed to encompass a wide range of scenarios. Each prompt within the ICE-Instruct model is responded to with a set of diverse answers, generated by sampling from the model's output distribution. This method ensures a comprehensive and varied dataset, essential for robust model training. The responses are further refined through a meticulous human annotation process, where experts rank them based on predefined criteria. This rigorous approach ensures the model is trained on high-quality, human-verified data, which is crucial for the model's ability to understand and apply complex information. More details and experimental comparsions are described in Section~\ref{sec:data_important}.

\noindent \textbf{Reward size Scaling:} 
In ICE-GRT, the scaling of the reward model is a critical factor in determining the overall effectiveness and efficiency of training. A larger reward model, denoted as \( R_\psi(s, a) \), where \( \psi \) represents the model parameters, is significant for several reasons.
Firstly, larger reward model can better capture complex environments and actions, essential in RLHF where the reward signal must accurately reflect human preferences and detailed task requirements.
Secondly, larger scale of reward size aids in generalizing across diverse prompts. This is vital for consistent performance in various scenarios, especially in ICE-GRT.

\noindent \textbf{KL-Control} is a crucial mechanism in PPO, especially when training with human feedback. 
A key aspect of KL-Control in this context is the regulation of divergence between the Actor and the Reference models. 
The KL divergence between these two models is monitored and controlled to ensure that the policy evolution adheres closely to the human feedback.
Moreover,  ICE-GRT training includes a clipping mechanism to avoid large, potentially destabilizing updates in the value function. This ensures that changes in the value function are moderate and accurately reflect real improvements as assessed by the Critic.
Furthermore, as an additional measure, KL Reward adjustment helps keep the actor model on the desired path as defined by human feedback. This aligns actor model updates more closely with human preferences.

\noindent \textbf{Advantage Normalization} enhances learning stability and efficiency in PPO-based RLHF. It adjusts the advantage estimates, making them more consistent and less variable. This is particularly beneficial in RLHF, where human feedback can introduce unpredictable variations. Normalizing the advantage helps the model to focus on the most relevant learning signals, leading to faster and more stable convergence.
The formula for Advantage Normalization is shown as follows:
$$\hat{A}_t^{\pi_{\theta}} = \frac{A_t^{\pi_{\theta}} - \mu_{A^{\pi_{\theta}}}}{\sigma_{A^{\pi_{\theta}}}},$$
where $\hat{A}_t^{\pi_{\theta}}$ represents the normalized advantage at time $t$, $A_t^{\pi_{\theta}}$ is the original advantage at time $t$, $\mu_{A^{\pi_{\theta}}}$ is the mean of the advantage, 
$\sigma_{A^{\pi_{\theta}}}$ is the standard deviation of the advantage.

\section{Experimental Details}

Our training process utilized the power of $64$ A$100$ GPUs, employing a multi-node, multi-GPU strategy to conduct ICE-GRT. 
Our models were trained and stored using the bf$16$ precision format. The learning rates were finely selected, with the actor learning rate set at $5e-6$ and the critic learning rate at $5e-7$. We maintained a clipping range of $0.2$. The discount factor $\gamma$ was kept constant at $0.95$, ensuring optimal balance in our training.
We are excited to announce the upcoming release and open-sourcing of our ICE-GRT $13$B model on Hugging Face, specifically tailored for scientific research purposes.

\subsection{Data Collection}
For our training corpus, we have crafted a novel mix of datasets. This includes a selection from publicly available resources, complemented by in-domain data. We have removed all the sensitive information, including usernames, email addresses, and personal details, to uphold the data privacy and security.
In essence, the dataset we have prepared for reward model and RLHF model is diverse and multi-faceted, covering a range of domains. It includes data relevant to public and domain-specific question-answering scenarios, as well as tasks involving multilingual data alignment. We generated $5$ distinct responses for every prompt in our data collection, utilizing our \textbf{ICE-Instruct} model. This process involves sampling from the model's output distribution, which guarantees a varied spectrum of answers. To optimally train our reward model, the data labelers carefully conducted manual labeling of the rankings for the $5$ distinct responses on 20,000 prompts. To enhance the human-annotation accuracy and reduce subjectivity among labelers, each prompt was independently evaluated by three labelers, establishing a thorough and reliable validation processverification process.

\begin{table*}[h!]
\centering
\tiny
\resizebox{\textwidth}{!}{
\begin{tabular}{|l|c|c|c|c|c|c|c|c|c|c|c|c|}
\hline
Model & \textbf{MMLU} & \textbf{AGIEval} & \textbf{BBH} & \textbf{AGIEval-ZH} & \textbf{ARC-E} & \textbf{ARC-C} & \textbf{HellaSWAG} & \textbf{Winogrande} & \textbf{RACE-M} & \textbf{RACE-H} & \textbf{GSM8K} & \textbf{Math} \\
 & \textbf{5-shot} & \textbf{few-shot} & \textbf{3-shot} & \textbf{few-shot} & \textbf{1-shot} & \textbf{1-shot} & \textbf{1-shot} & \textbf{1-shot} & \textbf{1-shot} & \textbf{1-shot} & \textbf{8-shot} & \textbf{4-shot} \\
\hline
LLaMA 7B & 24.66\% & 20.05\% & 33.48\% & 23.68\% & 30.01\% & 26.71\% & 24.58\% & 50.36\% & 26.74\% & 29.19\% & 13.80\% & 0.36\% \\
\hline
Llama2 7B & 40.91\% & 25.97\% & 38.21\% & 26.21\% & 62.37\% & 48.46\% & 25.39\% & 50.36\% & 45.75\% & 39.54\% & 17.51\% & 0.08\% \\
\hline
Vicuna 7B & 38.49\% & 22.71\% & 37.26\% & 27.00\% & 69.74\% & 46.33\% & 17.37\% & 49.80\% & 50.21\% & 46.83\% & 21.68\% & 0.96\% \\
\hline
ICE-Instruct 7B & 26.30\% & 15.95\% & 39.00\% & 31.14\% & 67.63\% & 45.31\% & 3.10\% & 36.07\% & 53.55\% & 52.09\% & 35.48\% & 0.82\% \\
\hline
LLaMA 13B & 38.42\% & 26.78\% & 38.28\% & 25.51\% & 67.63\% & 49.23\% & 28.90\% & 47.51\% & 52.23\% & 48.51\% & 18.42\% & 0.42\% \\
\hline
Llama2 13B & 49.57\% & 34.85\% & 45.89\% & 32.93\% & 76.52\% & 55.63\% & 37.17\% & 52.17\% & 57.73\% & 55.09\% & 28.66\% & 0.44\% \\
\hline
Vicuna 13B & 35.84\% & 28.68\% & 39.27\% & 30.33\% & 60.23\% & 40.96\% & 0.03\% & 5.84\% & 59.19\% & 60.69\% & 24.56\% & 0.66\% \\
\hline
ICE-Instruct 13B & 50.08\% & 24.51\% & 48.09\% & 34.15\% & 85.19\% & 66.89\% & 19.30\% & 47.99\% & 72.14\% & 56.52\% & 47.08\% & \textbf{1.02\%} \\
\hline
\textbf{ICE-GRT 13B} & \textbf{55.33\%} & \textbf{34.92\%} & \textbf{49.78\%} & \textbf{34.23\%} & \textbf{87.58\%} & \textbf{70.99\%} & \textbf{39.37\%} & \textbf{53.04\%} & \textbf{75.91\%} & \textbf{71.64\%} & \textbf{51.48\%} & 0.92\% \\
\hline
\hline
LLaMA 30B & 50.38\% & 34.87\% & 49.70\% & 30.68\% & 82.41\% & 60.67\% & 31.31\% & 51.30\% & 65.18\% & 64.18\% & 35.10\% & 0.58\% \\
\hline
Llama2-70B & 64.72\% & 43.99\% & 65.22\% & 39.52\% & 93.43\% & 79.61\% & 68.45\% & 69.69\% & 87.60\% & 85.13\% & 56.56\% & 3.72\% \\
\hline
\end{tabular}
}
\caption{Evaluating Benchmark Performance of Large Language Models in General Language Tasks.\label{tab:general_tasks_scores}}
\label{your-table-label}
\end{table*}

\subsection{General Task Evaluation}

Our evaluation of ICE-GRT using the GPT-Fathom framework~\cite{Zheng2023GPTFathomBL} focused on public general tasks. The objective was to benchmark ICE-GRT's performance against existing models and to understand its position in the landscape of current LLMs. We employed 12 benchmarks, which span across various capability categories such as language understanding, reasoning, etc. These benchmarks were carefully chosen to test a wide range of abilities, from basic language processing to complex problem-solving and decision-making tasks.
In our evaluation, we maintained alignment with the settings used in GPT-Fathom to ensure a fair and accurate comparison. This involved employing similar input formats, evaluation metrics, and environmental conditions.

\subsection{Manual Annotation-Based Evaluation}
\label{sec:manual_evaluation}

Our study incorporates a rigorous evaluation criteria, with a special emphasis on manual annotation for assessing the capabilities of LLMs, particularly in different applications. The criteria evaluates responses in $8$ essential categories, utilizing a scoring mechanism that prioritizes the most crucial aspects.

\noindent \textbf{Clarity}: Responses should be straightforward and precise, ensuring easy comprehension through specific, appropriate language.

\noindent \textbf{Accuracy}: The responses are expected to align closely with verified facts, as assessed by manual annotators. Actual fact can be validated.

\noindent \textbf{Completeness}: Evaluated for covering all aspects of the inquiry, providing comprehensive details for informed decision-making.

\noindent \textbf{Safety}: Focuses on ensuring no personal data is mishandled, with manual checks for data privacy.

\noindent \textbf{Courtesy}: Responses should be politically correct. e.g., gender identity, ethnic groups, etc.

\noindent \textbf{Comfortableness}: Responses must maintain a polite and respectful tone, containing inclusive vocabulary and reflect diversity at all times..

\noindent \textbf{Conciseness}: Emphasizes brevity in responses, without compromising on clarity or accuracy.

\noindent \textbf{Context}: Response must be related to the topic and relevant to the question.

Table~\ref{tab:manual_criteria} shows the weight and score of each categories to evaluate these criteria accurately, ensuring responses quality and relevance.

\begin{table}[htbp]
\centering
\scriptsize
\begin{tabular}{|l|c|c|c|c|}
\hline
Evaluation & Positive & Neutral & Negative & Weights \\
\hline
Clarity & 5 & 2 & 0 & 6 \\
\hline
Accuracy & 5 & 2 & 0 & 6 \\
\hline
Completeness & 5 & 2 & 0 & 6 \\
\hline
Safety & 5 & 2 & 0 & 3 \\
\hline
Courtesy & 5 & 2 & 0 & 3 \\
\hline
Comfortableness & 5 & 2 & 0 & 3 \\
\hline
Conciseness & 5 & 2 & 0 & 1 \\
\hline
Context & 5 & 2 & 0 & 1 \\
\hline
\end{tabular}
\caption{Manual Annotation-Based Evaluation Criteria. \label{tab:manual_criteria}}
\end{table}

\section{Results and Analysis}

\subsection{Results}

\textbf{Benckmarks Scores on General Tasks:}
Our analysis focuses on the performance of ICE-GRT $13$B, as compared to other models in similar and higher capacity categories.
As is shown in Table~\ref{tab:general_tasks_scores}, our ICE-GRT $13$B model demonstrates significant improvements over the LLaMa, Llama 2, Vicuna $13$B and LLaMa $30$B in both its pretrained and SFT across various general benchmarks, such as MMLU~\cite{hendryckstest2021}, AGIEval~\cite{Zhong2023AGIEvalAH}, BBH~\cite{srivastava2022beyond}, ARC~\cite{Xu2023LLMsAT}, HellaSWAG~\cite{Zellers2019HellaSwagCA}, RACE~\cite{Lai2017RACELR}, etc. It shows remarkable advancements in general language understanding and reasoning tasks, indicating enhanced comprehension and reasoning capabilities.
Remarkably, the ICE-GRT $13$B model has significantly narrowed the gap with the much larger Llama$2$ $70$B pretrain model. This comparison underscores the effectiveness of the ICE-GRT, compensating for smaller model size with more generalization capabilities.
The success of the ICE-GRT models suggests that the methodology, which likely includes components of human feedback and alignment, contributes significantly to the models' ability to understand and respond to complex prompts, a factor that is not solely dependent on model size.

\noindent \textbf{Human-Annotated Scores on In-Domain Task:}
In the in-domain evaluation presented in Table \ref{table:human_eval_table}, ICE-GRT distinctly outperforms Llama2 SFT $13$B and ICE-Instruct $13$B across several critical dimensions. Notably, ICE-GRT achieves the highest scores in clarity ($98.1\%$), accuracy (97.0\%), and completeness ($92.9\%$), underscoring its exceptional ability to deliver precise, comprehensive, and understandable responses. While it scores slightly lower in safety and comfort compared to its counterparts, it still maintains a high standard in these areas. The overall score of $95.5\%$ for ICE-GRT is a testament to its superior performance, significantly surpassing Llama2 SFT $13$B ($86.3\%$) and ICE-Instruct $13$B ($87.3\%$). This robust performance across multiple metrics confirms the introductory claims about ICE-GRT's capabilities, particularly in handling domain-specific tasks with a level of depth and precision not seen in current models.

\begin{table}[h!]
\centering \scriptsize
\begin{tabular}{|l|c|c|c|}
\hline
\textbf{} & \textbf{Llama2 sft} & \textbf{ICE-Instruct} & \textbf{ICE-GRT} \\
\hline
\textbf{Clarity} & 95.9\% & 88.5\% & \textbf{98.1\%} \\
\textbf{Accuracy} & 77.4\% & 84.44\% & \textbf{97.0\%} \\
\textbf{Completeness} & 64.8\% & 71.11\% & \textbf{92.9\%} \\
\textbf{Safety} & 96.6\% & \textbf{100\%} & 92.2\% \\
\textbf{Courtesy} & 100\% & 95.9\% & \textbf{100\%} \\
\textbf{Comfortable} & 96.6\% & \textbf{98.1\%} & 92.22\% \\
\textbf{Conciseness} & \textbf{95.1\%} & 93.33\% & 91.8\% \\
\textbf{Context} & \textbf{98.8\%} & 94.0\% & 98.1\% \\
\hline
\textbf{Overall Score} & 86.3\% & 87.3\% & \textbf{95.5\%} \\
\hline
\end{tabular}
\caption{Evaluating human-assessed scores for in-domain Large Language Models.}
\label{table:human_eval_table}
\end{table}

\subsection{Detailed Analysis}

\subsubsection{The importance of ICE-GRT Training Data}
\label{sec:data_important}

In the training of the ICE-GRT, we employed two distinct datasets for RLHF. The first dataset was uniquely produced by our ICE-Instruct model. For each prompt, five diverse responses were generated by sampling from the model outputs. These responses were then subjected to human annotation, where annotators ranked them according to predefined criteria. 
The second dataset originated from the GPT-4-LLM~ \cite{peng2023instruction}. It included ranked responses from GPT-4 and GPT-3.5, with the rankings automatically assessed by GPT-4.

\begin{figure}[h]
    \centering
    \begin{subfigure}[b]{0.23\textwidth}
        \centering
        \includegraphics[width=\textwidth]{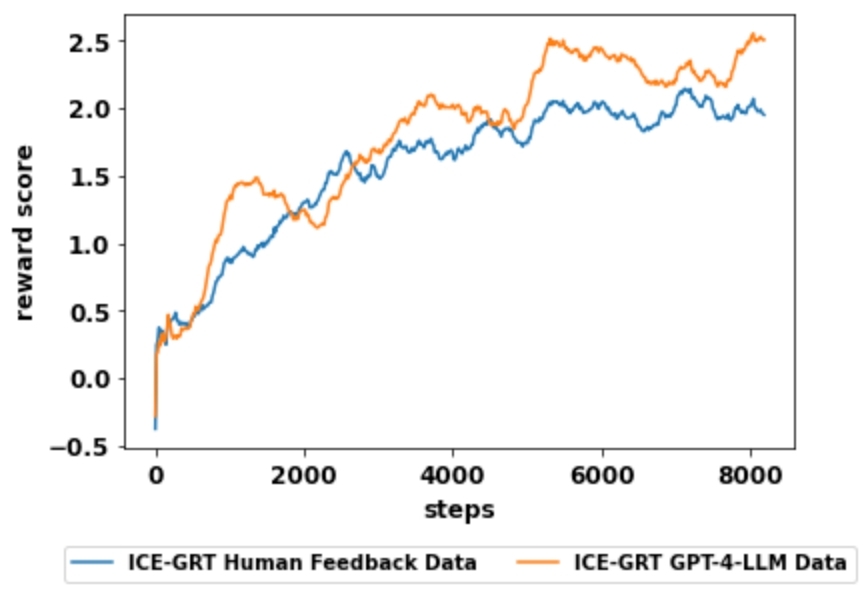}
        \caption{Reward score comparsion between different RLHF data.}
        \label{fig:data_ana1}
    \end{subfigure}
    \hfill
    \begin{subfigure}[b]{0.23\textwidth}
        \centering
        \includegraphics[width=\textwidth]{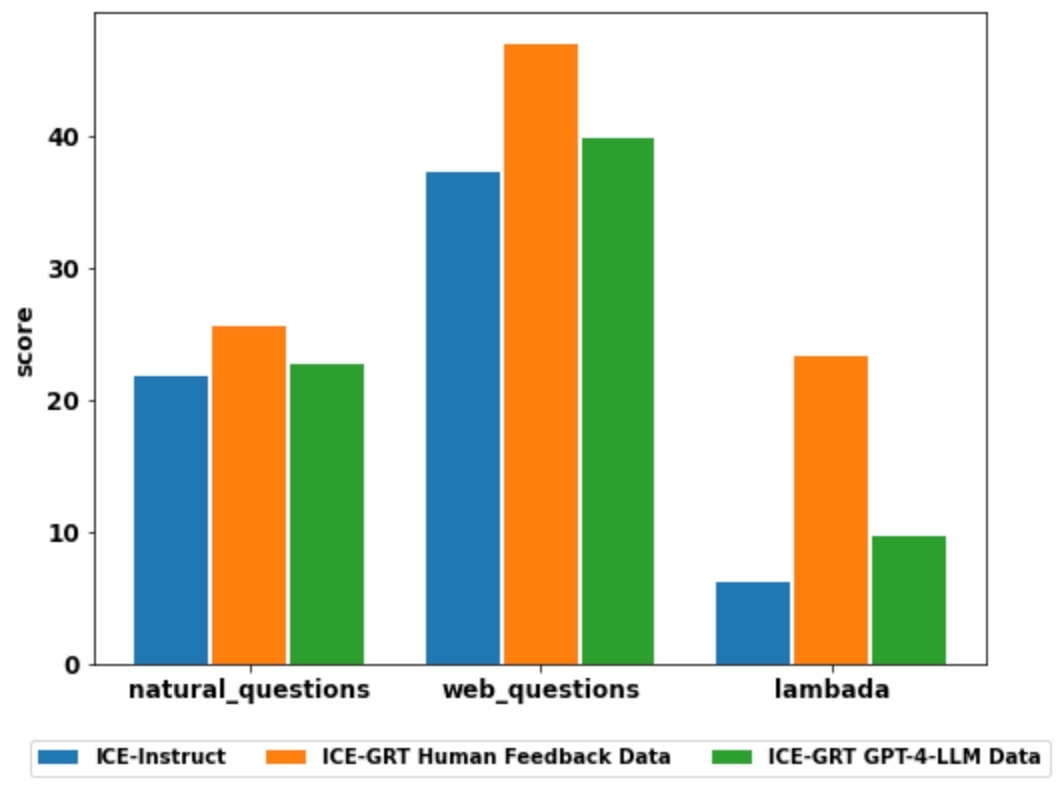}
        \caption{Benchmark performance between different models.}
        \label{fig:data_ana2}
    \end{subfigure}
    \caption{The influence of different training data.}
\end{figure}

Our findings reveal a significant performance disparity between models trained with these datasets, although we found that the reward score trends were similar during the ICE-GRT training shown in Figure~\ref{fig:data_ana1}. The ICE-GRT model, trained with our human-annotated dataset, demonstrated superior performance across general tasks and domain-specific tasks. As shown in Figure~\ref{fig:data_ana2}, on the Natural Question task, the ICE-GRT model outperformed ICE-Instruct by $4\%$. This gap increased to approximately $9.79\%$ on the Web Questions and $17.17\%$ on the LAMBADA benchmark. However, when we employed the GPT-4-LLM Dataset on ICE-GRT, we observe that the results were very close to those of ICE-Instruct, with only a $0.89\%$ increase in the Natural Questions.

A key aspect of ICE-GRT's success is its focus on `knowledge enhancement''. This process builds upon the ``knowledge mining'' during the ICE-Instruct, enabling the model to better align with human language preferences. 
This approach guarantees consistency and relevance in training data, which is crucial for the model to effectively build upon and evolve its existing knowledge. 
External data sources, despite their potential diversity, could not perfectly align with the model's knowledge structure.
The use of data generated by ICE-Instruct ensures a natural and effective enhancement of knowledge, as observed in ICE-GRT.

\begin{table*}[h!]
\centering \tiny
\begin{tabular}{ | m{2.5cm} | m{2.5cm} | m{2.5cm} | m{6cm} | }
\hline
\textbf{Question} & \textbf{ICE-Insturct $13$B} & \textbf{ICE-Insturct $33$B} & \textbf{ICE-GRT $13$B} \\ 
\hline
"this is a 100\% natural product for all females and all ages" what ad policy does the above text violate? & None, just make sure you're not lying about it. & The advertisement text may violate exaggerated claims policy. Overstates the product's effectiveness by claiming it is suitable for all females and all ages. & \textbf{The above text may violate exaggerated claims policy if the product is marketed as "100\% natural" without providing sufficient evidence or scientific backing for the claim}. Making an exaggerated or unsupported claim about a product's naturalness could mislead consumers and potentially violate advertising regulations. \textbf{To avoid violating this policy, advertisers should provide clear evidence to support any claims about a product's naturalness or provide a more nuanced description of its ingredients and production methods.} Additionally, advertisers should ensure that their claims are not overstated or misleading, and that they are consistent with any applicable laws and regulations governing product labeling and marketing. \\ 
\hline
Rewrite "this is a 100\% natural product for all females and all ages" to an ad without any violations & This product is suitable for all women and all age groups. & Experience the pure and natural goodness of this product, suitable for all ages and genders. & Introducing our natural product designed for all females and ages. \textbf{Our carefully crafted formulation is free of harmful chemicals and artificial additives, ensuring that you can use it with confidence.} Whether you're looking for a daily skincare routine or specialized care for a specific concern, our product is a safe and effective choice for all skin types. \textbf{Experience the benefits of natural ingredients and nourishing care with our product.} Suitable for all ages and skin types, our product is a safe and gentle choice for daily skincare. \\
\hline
\end{tabular}
\caption{Comparative Case Study in the Domain-Specific Moderation Task.}
\label{table:ad_policy}
\end{table*}

\subsubsection{Powerful ICE-GRT on General Task}
\begin{figure}[h]
    \centering
    \begin{subfigure}[b]{0.23\textwidth}
        \centering
        \includegraphics[width=\textwidth]{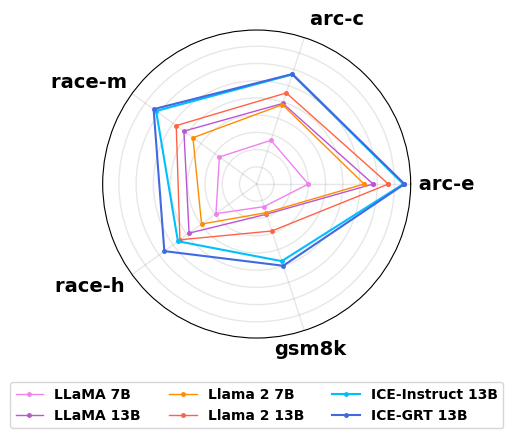}
        \caption{Language understanding and Reasoning Tasks.}
        \label{fig:general_ana1}
    \end{subfigure}
    \hfill
    \begin{subfigure}[b]{0.23\textwidth}
        \centering
        \includegraphics[width=\textwidth]{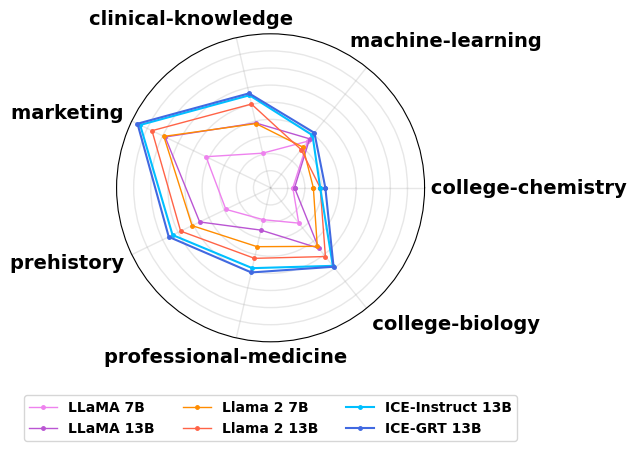}
        \caption{MMLU Task Score Comparsion.}
        \label{fig:general_ana2}
    \end{subfigure}
    \hfill
    \begin{subfigure}[b]{0.23\textwidth}
        \centering
        \includegraphics[width=\textwidth]{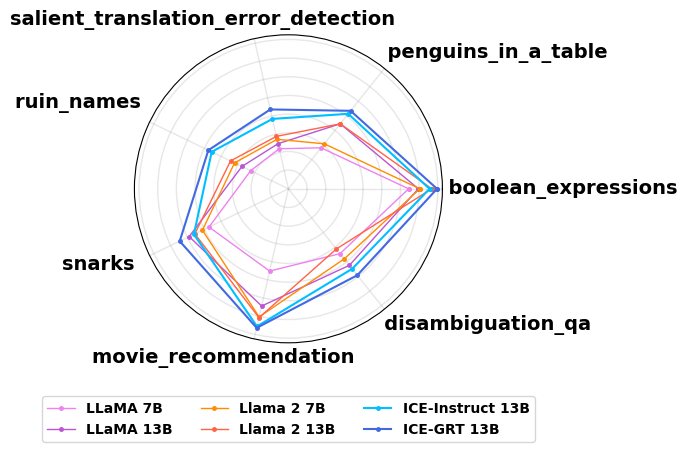}
        \caption{BBH Task Performance.}
        \label{fig:general_ana3}
    \end{subfigure}
    \hfill
    \begin{subfigure}[b]{0.23\textwidth}
        \centering
        \includegraphics[width=\textwidth]{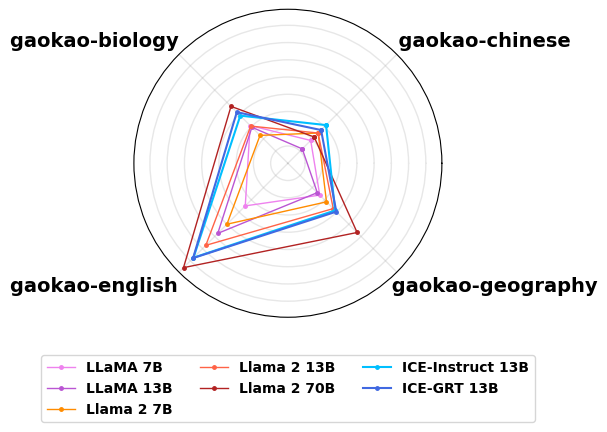}
        \caption{AGIEval-ZH Task.}
        \label{fig:general_ana4}
    \end{subfigure}
    \caption{Score Comparsions between different LLMs.}
    \label{fig:images}
\end{figure}

ICE-GRT model exhibits exceptional strength in tasks that are grounded in language understanding and reasoning. For instance, as shown in Figure~\ref{fig:general_ana1} and Figure~\ref{fig:general_ana3}, ICE-GRT $13$B demonstrates a robust ability in RACE, ARC, BBH, and GSM8K. Our model has achieved state-of-the-art performance in these tasks among models of the same size.

Moreover, the ICE-GRT $13$B model shows great performance in areas such as "marketing" and "college-biology", as depicted in Figure~\ref{fig:general_ana2}.
Remarkably, the ICE-GRT $13$B model surpasses the Llama 2 70B in several AGIEval benchmarks, most notably in tasks requiring language understanding and reasoning, such as "gaokao-chinese" as seen Figure~\ref{fig:general_ana4}. This outperformance stems from ICE-GRT knowledge enhanced contextual understanding and generative capabilities.

\subsubsection{The effectiveness of Advantage Norm}
\begin{figure}[h]
\centering
\includegraphics[width=0.4\textwidth,height=110pt]{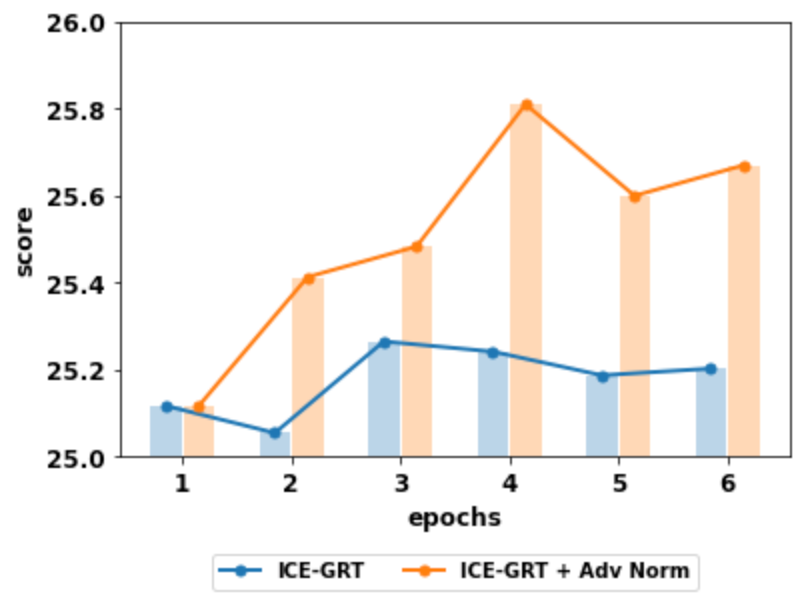}
\caption{Comparative Analysis of ICE-GRT and ICE-GRT Advantage Normalization on the Natural Question (NQ) Benchmark. The x-axis represents different epochs, while the y-axis shows the NQ scores. \label{fig:ana_adv}}
\end{figure}

The integration of Advantage Normalization and Reward Size Scaling significantly enhances ICE-GRT. These strategies contribute to improved training efficiency and better model performance, demonstrating their importance in the context of RLHF. Applying Advantage Normalization, which stabilizes learning by normalizing advantage estimates, led to improvement in Natural Question benchmark over ICE-GRT baseline. As shown in Figure~\ref{fig:ana_adv},this strategy is crucial for enhancing the model's sensitivity to the subtleties of human feedback, leading to more effective learning outcomes.

\subsection{Case Study on Domain-Specific Task}

We provide a comparative analysis of the responses generated by different models, specifically ICE-Instruct $13$B, $33$B, and ICE-GRT $13$B, revealing varying levels of sensitivity and creativity in addressing advertising policy adherence and rewriting for compliance. As is shown in Table~\ref{table:ad_policy}, while ICE-Instruct $13$B takes a more direct and less cautious approach, ICE-Instruct $33$B and ICE-GRT $13$B demonstrate a progressive increase in policy awareness and creative compliance. 

ICE-GRT, in particular, shows a comprehensive understanding of advertising regulations and the importance of substantiated claims, reflecting its advanced capability in nuanced and responsible communication. In the first case, ICE-GRT displayed the highest sensitivity to policy adherence, highlighting the risk of violating exaggerated claims policy, especially if the product is marketed as "100\% natural" without adequate evidence. It emphasizes the need for evidence-based advertising and compliance with regulations. In the second case, ICE-GRT Provided the most detailed and cautious rewrite, ensuring compliance with advertising policies. It focuses on natural ingredients, absence of harmful chemicals, and suitability for all females and ages, while avoiding exaggerated claims.

In this section, we have showcased only a small fraction of our model's capabilities, focusing primarily on the in-domain task of ad moderation. However, the scope of our model, ICE-GRT, extends far beyond this singular function. Within the appendices, we demonstrate its proficiency across a myriad of domain-specific tasks. These include, but are not limited to, Poem Generation, Text-to-Table, Multiple Round Dialogue~(Appendix~\ref{append:1}), Chemistry Response Generation~(Appendix~\ref{append:2}), Code Generation~(Appendix~\ref{append:3}), Ads Text Generation, Labeling Text~(Appendix~\ref{append:4}), and Multi-lingual Response~(Appendix~\ref{append:5}), etc. The choice of the acronym "ICE" for ICE-GRT is deliberate. It represents our aspiration to catalyze an 'ice-breaking' moment in the research of LLMs. This reflects our hope to encourage researchers to explore and realize the broad possibilities of ICE-GRT in a range of tasks. We aim to pave the way for novel discoveries and advancements in the field, demonstrating that the capabilities of our model are as extensive and varied as they are innovative.

\section{Conclusion}
ICE-GRT model represents a significant leap forward in the realm of LLMs, particularly in enhancing domain-specific performance. Leveraging the principles of Reinforcement Learning from Human Feedback, ICE-GRT demonstrates exceptional capabilities in both general and in-domain tasks, outperforming standard models in accuracy and depth. Moreover, our model have strong ability to generate detailed analyses of the reasons behind the answer. Our research uncovers several aspects of RLHF, providing insights into effective training methodologies and highlighting the importance of factors like Appropriate Data, Reward Size Scaling, KL-Control, etc. ICE-GRT's training phases, including knowledge learning, mining, and enhancement, contribute to its advanced abilities in aligning with human preferences. We hope that ICE-GRT will accelerate the ``ice-breaking'' process in LLM research, encouraging further exploration.

\section*{Acknowledgements}
We deeply appreciate Youlong Cheng, Guokun Lai, Yingtong Bu, Zheng Zhang, Fan Qu for their help at the early stage of this project. Moreover, we convey appreciation to Hanzhi Zhou, Yijie Zhu, Xuan Zou for their engineering support to build key components of the infrastructure. We extend our gratitude to Hang Wu, Ruoqi Zhang and Ruohong Zhang for their insightful discussions that contributed to this paper. Furthermore, we thank anonymous reviewers for their valuable suggestions.

\bibliography{acl2024}



\newpage
\onecolumn
\appendix

\section{Poem Generation, Text-to-Table, Multiple Round Dialogue.}
\label{append:1}

This appendix section presents the multifaceted capabilities of the ICE-GRT, as it navigates through three distinct tasks: Poem Generation, Text-to-Table, and Multiple Round Dialogue. Each section showcases a unique aspect of the ICE-GRT prowess. In Poem Generation, we explore its creative capacity to craft expressive content. The Text-to-Table task demonstrates its proficiency in data structuring and presentation. Finally, the Multiple Round Dialogue section exhibits its skill in engaging in predictive and analytical discourse. These tasks collectively underscore strong abilities of the ICE-GRT, exemplifying its potential in diverse linguistic and analytical applications.

\begin{table*}[h!]
\centering \tiny
\begin{tabular}{  m{4cm} | m{10cm} }
\hline
\textbf{Questions} & \textbf{Responses} \\
\hline
Compose a poem to analyze \textbf{American economy} with a positive attitude. & 
The American economy is strong and vibrant,

A shining beacon of hope and progress.

From coast to coast, businesses thriving,

Innovation and prosperity on the rise.

\vskip 0.5cm

From the tech giants in Silicon Valley,

To the small startups in every town,

The American spirit of enterprise,

Driving growth and creating jobs galore.

\vskip 0.5cm

From Wall Street to Main Street,

Investors and consumers alike,

Confidence in the future is high,

As the economy continues to expand and thrive.

\vskip 0.5cm

With a positive attitude and a can-do spirit,

The American economy will continue to grow and prosper,

Bringing opportunity and prosperity,

To all who work hard and strive for success. \\

\hline
\end{tabular}
\caption{Case Study in Poem Generation task.}
\label{table:poem}
\end{table*}

\begin{table*}[h!]
\centering \tiny
\begin{tabular}{  m{4cm} | m{10cm} }
\hline
\textbf{Questions} & \textbf{Responses} \\
\hline
Then Generate a table to show \textbf{Americans main economic} metrics, such as GPP, PPP, CPI, PPI, etc &

Here is a table showing some of the main economic metrics for the United States:

\includegraphics[width=6cm]{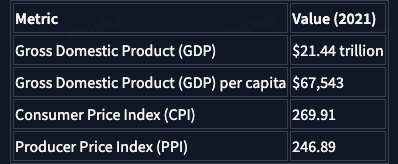}

Note: These values are subject to change and may not reflect the most up-to-date information. It's important to consult the latest official sources for the most current economic data.

 \\

\hline
\end{tabular}
\caption{Case Study in Text-to-Table Task.}
\label{table:text_to_table}
\end{table*}

\begin{table*}[h!]
\centering \tiny
\begin{tabular}{  m{4cm} | m{10cm} }
\hline
\textbf{Questions} & \textbf{Responses} \\
\hline

Predict \textbf{American economy} next year based on the data above and give your reasons. & 
As an AI language model, I cannot predict the future with certainty, but I can provide some insights based on the economic data available as of my knowledge cutoff date (September 2021). The United States has experienced a strong economic recovery in 2021, with GDP growth projected to be around 6\% for the year. This growth has been driven by a combination of factors, including strong consumer spending, increased business investment, and a rebound in international trade. The Consumer Price Index (CPI) has also risen steadily throughout 2021, reflecting higher inflationary pressures in the economy.
\vskip 0.5cm
Looking ahead to the next year (2022), there are several factors that may impact the U.S. economy. One key factor to watch will be the trajectory of the COVID-19 pandemic, as the ongoing pandemic continues to pose risks to economic growth and stability. Additionally, the U.S. Federal Reserve has signaled that it may begin tapering its asset purchases in the coming months, which could impact financial markets and interest rates.
\vskip 0.5cm
Assuming a continued economic recovery and a gradual normalization of monetary and fiscal policies, it is possible that the U.S. economy could experience modest growth in 2022. However, there are also risks and uncertainties that could impact the outlook, including the potential for new waves of the pandemic, geopolitical risks, and ongoing supply chain disruptions. Overall, the outlook for the U.S. economy in 2022 will depend on a variety of factors, and it is important for policymakers and businesses to remain vigilant and adaptable in the face of evolving economic conditions.

 \\

\hline
\end{tabular}
\caption{Case Study in Multiple Round Dialogue task.}
\label{table:poem_multiple_round}
\end{table*}

\newpage
\onecolumn

\section{Chemistry Response Generation.}
\label{append:2}

\begin{table*}[h!]
\centering \tiny
\begin{tabular}{  m{4cm} | m{10cm} }
\hline
\textbf{Questions} & \textbf{Responses} \\
\hline

Help me with the chemistry equations, such as H2O related equations.
 & 
Here are some commonly used chemical equations related to H2O:

1. Combustion of hydrogen gas (H2):

2H2 + O2 -> 2H2O
\vskip 0.5cm
2. Decomposition of water (H2O) into hydrogen gas (H2) and oxygen gas (O2) using electricity:

2H2O -> 2H2 + O2
\vskip 0.5cm
3. Reaction of hydrogen gas (H2) and oxygen gas (O2) to form water (H2O):

2H2 + O2 -> 2H2O
\vskip 0.5cm
These equations are just a few examples of how H2O is involved in various chemical reactions. There are many other equations and reactions that involve H2O, depending on the specific context and conditions. It's important to understand the underlying chemistry and physics of these equations and reactions in order to fully comprehend their implications and applications. Additionally, it's important to follow proper safety protocols and procedures when handling chemicals and performing experiments, as some reactions can be hazardous or even dangerous if not properly controlled and monitored.

 \\

\hline
\end{tabular}
\caption{Case Study in Chemistry Generation Task.}
\label{table:chemistry}
\end{table*}

\section{Code Generation.}
\label{append:3}

\begin{table*}[h!]
\centering \small
\begin{tabular}{  m{4cm} | m{10cm} }
\hline
\textbf{Questions} & \textbf{Responses} \\
\hline

Write a function to count the appearing times of each word, then write another function to sort this list, but you cannot use any libs in python.
 & 
 Sure, here's a Python function that counts the occurrences of each word in a given string and returns a dictionary with the word as the key and its count as the value:
\begin{lstlisting}[language=Python]
def count_words(text):
    words = text.lower().split()
    word_counts = {}
    for word in words:
        if word in word_counts:
            word_counts[word] += 1
        else:
            word_counts[word] = 1
    return word_counts
\end{lstlisting}
To sort the dictionary of word counts, you can use the sorted() function along with a lambda function that compares the values of the dictionary items:
\begin{lstlisting}[language=Python]
def sort_word_counts(word_counts):
    sorted_counts = sorted(word_counts.items(), 
                key=lambda x: x[1], reverse=True)
    return sorted_counts
\end{lstlisting}
Here's how you can use these functions together to count the occurrences of each word in a string and sort the word counts:
\begin{lstlisting}[language=Python]
text = "The quick brown fox jumped over the lazy dog. The dog barked at the fox."
word_counts = count_words(text)
sorted_word_counts = sort_word_counts(word_counts)
print(sorted_word_counts)
\end{lstlisting}
This will output:
\begin{lstlisting}
[('fox', 2), ('dog', 2), ('the', 2), ('quick', 1), ('brown', 1), ('jumped', 1), ('over', 1), ('lazy', 1), ('barked', 1)]
\end{lstlisting}
\\

\hline

\end{tabular}
\caption{Case Study 1 in Code Generation Task.}
\label{table:code1}
\end{table*}

\begin{table*}[h!]
\centering \small
\begin{tabular}{  m{4cm} | m{10cm} }
\hline
\textbf{Questions} & \textbf{Responses} \\
\hline

By using the words above as corpus, try to implement word2vec code based on its theory &
Sure, here's an implementation of word2vec in Python based on the theory described in the paper "Word2Vec: Negative Sampling for Fast and Accurate Learning of Word Vectors" by Mikolov et al. This implementation uses the Skip-Gram algorithm to train a neural network to predict nearby words given a target word, and vice versa. The resulting word vectors can be used for various natural language processing tasks, such as text classification, language translation, and sentiment analysis.
\vskip 0.5cm
Here's the implementation:
\begin{lstlisting}[language=Python]
import numpy as np
import gensim
from gensim.models import CoherenceModel
from gensim.models.coherencemodel import CoherenceModel as Coherence
from gensim.models.word2vec import Word2Vec
from gensim.corpora import Dictionary
from gensim.models.ldamodel import LdaModel

# Set up the corpus and dictionary
text = "The quick brown fox jumped over the lazy dog. The dog barked at the fox.".lower()
tokens = [word for word in text.split()]
dictionary = Dictionary(tokens)

# Train the word vectors using the Skip-Gram algorithm
model = Word2Vec([text], min_count=1, size=100, window=5, workers=4)

# Evaluate the word embeddings using coherence analysis
model_coherence = Coherence(model.wv, model.wv, dictionary=dictionary, coherence='c_v', coherence_mode='vector')
coherence_score = coherence_model.get_coherence()
print(f"Coherence score: {coherence_score}")

# Train a LDA model on the word embeddings to identify topics in the corpus
lda_model = LdaModel(corpus=[model.wv], id2word=dictionary, num_topics=5, random_state=42)
topics = lda_model.print_topic_words(num_words=5)
print(topics)
\end{lstlisting}
This implementation uses the gensim library, which provides an easy-to-use interface for training word embeddings and other natural language processing models. The Word2Vec class is used to train the word vectors using the Skip-Gram algorithm, while the CoherenceModel class is used to evaluate the word embeddings using coherence analysis. Finally, the LdaModel class is used to train a LDA model on the word embeddings to identify topics in the corpus. The resulting topics can provide insights into the main themes and topics present in the corpus.
\\
\hline

\hline
\end{tabular}
\caption{Case Study 2 in Code Generation Task.}
\label{table:code2}
\end{table*}

\newpage
\onecolumn

\section{Ads text Generation, Label text, Multiple Round Dialogue.}
\label{append:4}

\begin{table*}[h!]
\centering \small
\begin{tabular}{  m{4cm} | m{10cm} }
\hline
\textbf{Questions} & \textbf{Responses} \\
\hline

I want to post decoration ads, please generate some candidate for me to promote my devices and materials. & 

Sure, here are some sample ad copy ideas for decoration products:

1. "Transform your home into a work of art with our high-quality decorative items. From wall art and sculptures to accent pieces and lighting, we've got everything you need to make your space truly unique."

2. "Unleash your creativity and elevate your home decor with our wide selection of decorative materials. Choose from a variety of textures, colors, and patterns to create a personalized look that's all your own."

3. "Looking for the perfect finishing touch for your home decor? Look no further than our selection of decorative accents. From trendy tapestries and macrame wall hangings to stylish vases and planters, we've got everything you need to add the finishing touch to your space."

4. "Create a warm and inviting atmosphere in your home with our selection of decorative lighting. From chic pendant lights and table lamps to statement chandeliers and wall sconces, we have the perfect lighting solution for every room in your home."

5. "Make your home a haven with our collection of decorative cushions and throws. Choose from a variety of patterns and colors to add a pop of personality to your sofa, armchair, or bed, or layer multiple cushions for a cozy and inviting look."

I hope these ad copy ideas help inspire your decoration ads!
\\

\hline

Transform the ad texts above into semantically negative
 & 
Sure, here are some sample ad copy ideas for decoration products with a negative spin:

1. "Don't settle for mediocre home decor - upgrade to our high-quality items and still end up with a bland, uninspired look."

2. "Looking for a way to make your home decor even more generic and unoriginal? Our selection of decorative materials is just what you need!"

3. "Want to make your home decor look even more outdated and unstylish? Our selection of decorative accents will do the trick!"

4. "Create an even duller and more uninviting atmosphere in your home with our selection of decorative lighting. Choose from a variety of boring and unflattering options to make your home even less welcoming."

5. "Transform your home into a cold, impersonal space with our collection of decorative cushions and throws. Choose from a variety of lifeless and uninspired patterns and colors to make your sofa, armchair, or bed even less comfortable and inviting."

Note: It's generally not recommended to use negative or sarcastic language in advertising as it may turn off potential customers and harm the brand's reputation. It's usually more effective to use positive and aspirational language that appeals to the customer's desires and values.
\\

\hline

Assume you are a labeler, please label each texts above into "Negative", "Positive", "Neutral"
 & 
Sure, here are the labels for each of the ad copy ideas:

1. Negative

2. Negative

3. Negative

4. Negative

5. Positive

Note: The labels are subjective and based on the overall tone and language used in the ad copy ideas. The same ad copy idea could potentially be labeled differently by different labelers depending on their personal interpretation and perspective.
\\

\hline
We have one more requirement, when labeling the texts, please also add probabilities of the label.
 & 
 Sure, here are the labels and probabilities for each of the ad copy ideas:
1. Negative (95\%)

2. Negative (80\%)

3. Negative (85\%)

4. Negative (70\%)

5. Positive (60\%)

Note: The probabilities are based on my analysis of the overall tone and language used in the ad copy ideas, and are intended as an estimate rather than a definitive measurement. The same ad copy idea could potentially have different probabilities assigned by different labelers depending on their personal interpretation and perspective. Additionally, the probabilities are subject to variation based on the specific context and audience being targeted, as well as any changes or updates made to the ad copy ideas over time.

 \\

\hline
\end{tabular}
\caption{Case Study in Ads related text Generation and Labeling Task.}
\label{table:ad_text_case}
\end{table*}

\newpage
\onecolumn

\section{Multi-lingual Response}
\label{append:5}

\begin{table*}[h!]
\centering \small
\begin{tabular}{  m{4cm} | m{10cm} }
\hline
\textbf{Questions} & \textbf{Responses} \\
\hline
\begin{CJK}{UTF8}{gbsn}
我很喜欢周杰伦的《简单爱》这首歌曲。可以帮我分析一下这首歌吗？
\end{CJK}
 & 
\begin{CJK}{UTF8}{gbsn}
当然可以。周杰伦的《简单爱》这首歌是一首抒情的流行歌曲，以简单的旋律和易于记忆的歌词著称。以下是对这首歌的一些分析：

1.歌词：歌曲的歌词简单易懂，表达了简单的爱情，没有太多复杂的情感和抽象的比喻。歌词中的“简单爱”是指一种真实、纯粹的爱情，没有太多的承诺和纷乱的感情，而是一种简单而又深刻的感情体验。

2.旋律：这首歌的旋律非常易于记忆，很快就能让人唱出来。旋律简单而优美，具有一种轻快的节奏感，让人感到快乐和舒适。

3.编曲：这首歌的编曲非常简单，主要由钢琴、吉他和弦乐组成。钢琴和吉他的和声营造了一种温馨的氛围，弦乐则增加了歌曲的感染力和气氛。

4.表现手法：这首歌的表现手法非常简单，主要是周杰伦的嗓音以及他的情感投入。周杰伦在演唱这首歌时，通过自然而然的呼吸、声调的变化和情感的表达，让人感到他真正地体验到了简单的爱情，并且让人产生共鸣。

综上所述，《简单爱》这首歌是一首非常易于记忆、简单易懂、温馨感人的流行歌曲，通过周杰伦的演唱和简单的编曲，传递出了一种简单而真实的爱情感受。
\end{CJK}
\\
\hline
\begin{CJK}{UTF8}{gbsn}
describe the panda in Chinese
\end{CJK}
 & 
\begin{CJK}{UTF8}{gbsn}
熊猫（xiong mao）是一种来自中国的熊科动物，也被称为大熊猫。它们有黑白相间的柔软皮毛和圆圆的脸部，是世界上最受欢迎的动物之一。熊猫主要生活在中国的山区，以竹子为食，是濒危物种之一。它们是中国文化的重要象征，也被广泛认为是保护自然环境的代表。
\end{CJK}
\\

\hline
\end{tabular}
\caption{Case Study in Multi-lingual Response Task.}
\label{table:multi_linguistic}
\end{table*}

\newpage
\onecolumn

\section{Sampled Questions in In-domain task.}

\begin{table*}[h!]
\centering \small
\begin{tabular}{  m{0.5cm}  m{13cm} }
\hline
\textbf{No.} & \textbf{Questions} \\
\hline
1 & What is an auction ad? \\
2 & What is pre-audit and smart fix in ads platform? \\
3 & What is machine moderation \\
4 & What is human moderation in advertising \\
5 & Will the company use machine learning models to moderate ad groups? \\
6 & Explain Advertiser Experience (AdEx) \\
7 & Generate a list of advertising metrics on platform \\
8 & How to advertise on platform \\
9 & How to create an ad \\
10 & How can I create an ad group \\
11 & Can I add a link to my account in an ad? \\
12 & Can I edit a bid? \\
13 & Where can I see my account's transaction? \\
14 & Why were my ad groups rejected? \\
15 & Why did you reject my ads? \\
16 & How do I fix an ad that was disapproved in different locations? \\
17 & What ad policy does the following text violate? Lonely and thirsty for a man \\
18 & What ad policy does the following text violate? This Streamline Makes Your Cheeky Hidden Hips Look Like a Juicy Peach \\
19 & What ad policy does the following text violate? Red Boost is a powerful new formula for boosting male sexual health, weight loss, muscle building. \\
20 & Generate 10 suggestions to post non-political content ads on platform \\
21 & show me the details of prohibited industry policy in the united States \\
22 & Classify the following comment into labels [non-delivery, description doesn't match item, data theft, body harm, unreasonable expense]: "Didn't get it!" \\
23 & Classify the following text into these ad policies [political content policy, sexual hint policy, no violated policy]. "How are you" \\
24 & List some examples concerning adult products, which is prohibited \\
25 & List some examples concerning sexual hint, which is restricted \\
26 & List some reasons why sexual hint is not allowed on platform \\
27 & Generate 10 reasons why sexual hint is not allowed on platform \\
28 & "This is a 100\% natural product for all females and all ages" what ad policy does the above text violate? (A) no violations (B) exaggerated (C) political content (D) absolute terms (E) none of the above \\
29 & "This is a 100\% natural product for all females and all ages" what ad policy does the above text violate? \\
30 & Explain the reasons why the following text violates exaggerated description ad policy. "This is a 100\% natural product for all females and all ages" \\
31 & Rewrite "This is a 100\% natural product for all females and all ages" to an ad without any violations \\
32 & Rewrite "Number 1 product in the world" to the text that doesn't violate exaggerated description policy \\
33 & Generate a list of 10 common reasons why ad groups are rejected \\
34 & How to lose weight without effort, write more than 1000 words \\
\hline
\end{tabular}
\caption{Sampled Questions in Ad moderation Task.}
\label{table:ad_policy_samples}
\end{table*}


\end{document}